\documentclass[10pt,twocolumn]{article}

\newif\ifblind

\usepackage[margin=0.75in]{geometry}
\usepackage{times}
\usepackage{microtype}
\usepackage{graphicx}
\usepackage{amsmath,amssymb}
\usepackage{authblk}
\usepackage{caption}
\usepackage{enumitem}
\usepackage{xcolor}
\usepackage{hyperref}
\usepackage{adjustbox}
\hypersetup{hidelinks}

\usepackage[compact]{titlesec}
\titlespacing*{\section}{0pt}{0.3em}{0.3em}
\title{Regime Mapping of Oscillatory States in Balanced Spiking Networks with Multiple Time Scales}

\author{
Tsung-Han Kuo$^{1}$, Tzu-Chia Tung$^{2}$\\
\small $^{1}$Graduate Institute of Networking and Multimedia, National Taiwan University\\
\small $^{2}$Graduate Institute of Vehicle Engineering, National Changhua University of Education\\
\texttt{d05944019@ntu.edu.tw, tct@cc.ncue.edu.tw}
}
\date{}

\begin{document}


%

\maketitle
\begin{abstract}
Balanced spiking networks can transition between silent, asynchronous-irregular, and oscillatory states depending on interacting synaptic and temporal time scales, while their joint parameter structure remains incompletely characterized. In this work, we systematically map how postsynaptic decay (\(\tau_s\)), conduction delay (\(d\)), and plasticity rate (\(\lambda_p\)) jointly shape oscillatory regimes in recurrent leaky integrate-and-fire networks. By combining Brian2 simulations across the \((\tau_s,d,\lambda_p)\) space with a coarse Hopf-reference boundary, we construct regime maps that directly visualize SIL--AI--OSC transitions and corresponding spectral prominence landscapes. The mapped results show that increasing \(\lambda_p\) expands oscillatory regions toward shorter \(\tau_s\) and moderate-to-long delays, while prominence maps identify parameter regions with the strongest rhythmic coherence. Representative control experiments further connect this global landscape to local rhythm-forming mechanisms, showing that STDP freezing weakens rhythmic coherence whereas delay jitter enhances it with minimal change in mean firing rate. As a result, these findings provide a useful reference for operating-point selection, synchrony modulation studies, and future biologically grounded spiking-network modeling within similar balanced-network settings.
\end{abstract}
%
%

\vspace{0.4em}

\section*{Introduction}
Balanced excitation and inhibition (E/I) allows cortical circuits to exhibit a range of collective dynamics, from asynchronous irregular firing to coherent oscillations~\cite{Brunel2000,Vogels2011}.
%
%
Changes in synaptic coupling, temporal integration, or transmission delay can shift recurrent networks across these dynamical states, affecting rhythm generation and large-scale synchrony~\cite{ErmentroutTerman2010,Morrison2008}.  
%
%
Classical balanced-network studies have long described these collective behaviors~\cite{Brunel2000}, while later theoretical and computational work showed that spike-timing-dependent plasticity (STDP) further modulates stability and synchronization in recurrent spiking systems~\cite{Morrison2008,Roxin2004}. 
%

Previous studies have approached network dynamics from complementary perspectives, including delay-driven oscillations~\cite{Roxin2004,Buzsaki2006}, multi-timescale plasticity~\cite{Zenke2017}, and concurrent stability and learning via fast and slow synaptic adaptation~\cite{NatureComms2024}. More recently, work has suggested that multiple synaptic time scales, including fast postsynaptic current integration dynamics and slower plasticity processes, help recurrent networks remain both adaptive and stable~\cite{NatureComms2024}.
%

However, the combined effects of postsynaptic integration (\(\tau_s\)), conduction delay (\(d\)), and plasticity rate (\(\lambda_p\)) on oscillatory states have not been systematically mapped. To address this gap, our regime maps directly visualize how these interacting time scales jointly shape oscillatory regimes and drive transitions between asynchronous-irregular (AI) and oscillatory (OSC) states in balanced spiking networks.
%
%

Here, we systematically map the parameter space of recurrent leaky integrate-and-fire (LIF) networks to examine how \(\tau_s\), \(d\), and \(\lambda_p\) determine oscillatory states.  
%
%
Using Brian2 simulations~\cite{Stimberg2019} together with a linearized Hopf-boundary approximation as a coarse theoretical reference, we construct regime maps that identify parameter regions supporting oscillatory activity.
%
%

\section*{Methods}
We simulated a recurrent spiking network consisting of 160 excitatory and 40 inhibitory leaky integrate-and-fire (LIF) neurons with sparse random connectivity (\(p = 0.1\)).  
%
%
Synaptic inputs were modeled as exponential postsynaptic currents with decay constant (\(\tau_s\)), which determines the effective postsynaptic integration window, together with a conduction delay parameter (\(d\)).
%
%
Excitatory-to-excitatory synapses followed a pair-based STDP rule with plasticity rate \(\lambda_p\), while all other synapses remained static.  
%

We systematically varied \(\tau_s \in [5, 30]~\mathrm{ms}\), \(d \in [0, 10]~\mathrm{ms}\), and \(\lambda_p \in \{0, 5\times10^{-4}, 2\times10^{-3}\}\) to construct regime maps of their joint effects on network activity. The \(\tau_s\) and \(d\) ranges were centered on commonly used cortical-scale physiological time constants with limited extension to better resolve AI--OSC transitions. 
The three \(\lambda_p\) slices were chosen from preliminary coarse sweeps over increasing plasticity strengths to capture representative low, intermediate, and high plasticity regimes.
%

Neuron and synaptic dynamics follow standard LIF and exponential-synapse equations, where \(\tau_m\) and \(\tau_s\) denote the membrane and synaptic decay time constants:
%
%
\begin{align}
\tau_m \frac{dv_i}{dt} &= v_{\mathrm{rest}} - v_i + I_{\mathrm{syn},i} + I_{\mathrm{ext},i}, \\
%
\tau_s \frac{dI_{\mathrm{syn},i}}{dt} &= -I_{\mathrm{syn},i} + \sum_j w_{ij}s_j(t - d_{ij}).
\end{align}
%
Excitatory synapses adapted according to a pair-based STDP rule scaled by \(\lambda_p\):
\begin{equation}
\Delta w_{ij} = \lambda_p \big[ A_+ e^{-\Delta t / \tau_+} - A_- e^{\Delta t / \tau_-} \big].
\end{equation}
Here, \(\Delta t = t_{\mathrm{post}} - t_{\mathrm{pre}}\) denotes the post--pre spike timing difference.
%
%

Population activity was measured from smoothed firing-rate traces (5~ms Gaussian kernel) after excluding a 0.5~s burn-in period from the total 8~s simulation.  
%
%

To interpret the transition between asynchronous and oscillatory regimes, a linearized rate-model Hopf approximation \(H(\tau_s,d)\) served as a coarse theoretical reference. From a dynamical-systems perspective, Hopf bifurcation analysis is commonly used to estimate when irregular fluctuations transition into sustained rhythmic activity. In this work, the resulting Hopf-like boundary serves as a visual theoretical guide for identifying where the simulated AI region begins to enter OSC regimes.
%
%
Complementing this boundary analysis, oscillatory strength was quantified using PSD prominence, defined as the peak-to-median ratio within 5--100 Hz, where \(f_0\) was defined as the dominant spectral peak within 8--70 Hz. This metric was chosen to emphasize rhythmic coherence independently of mean firing-rate changes.
%
%

All conditions were repeated across five random seeds, and summary statistics are reported as mean~\(\pm\)~s.d.\ for firing rate, \(f_0\), and PSD prominence.
%

\section*{Results and Discussion}
\begin{figure}[t]
  \centering
  \vspace{-0.3em}
  \includegraphics[width=1.0\linewidth]{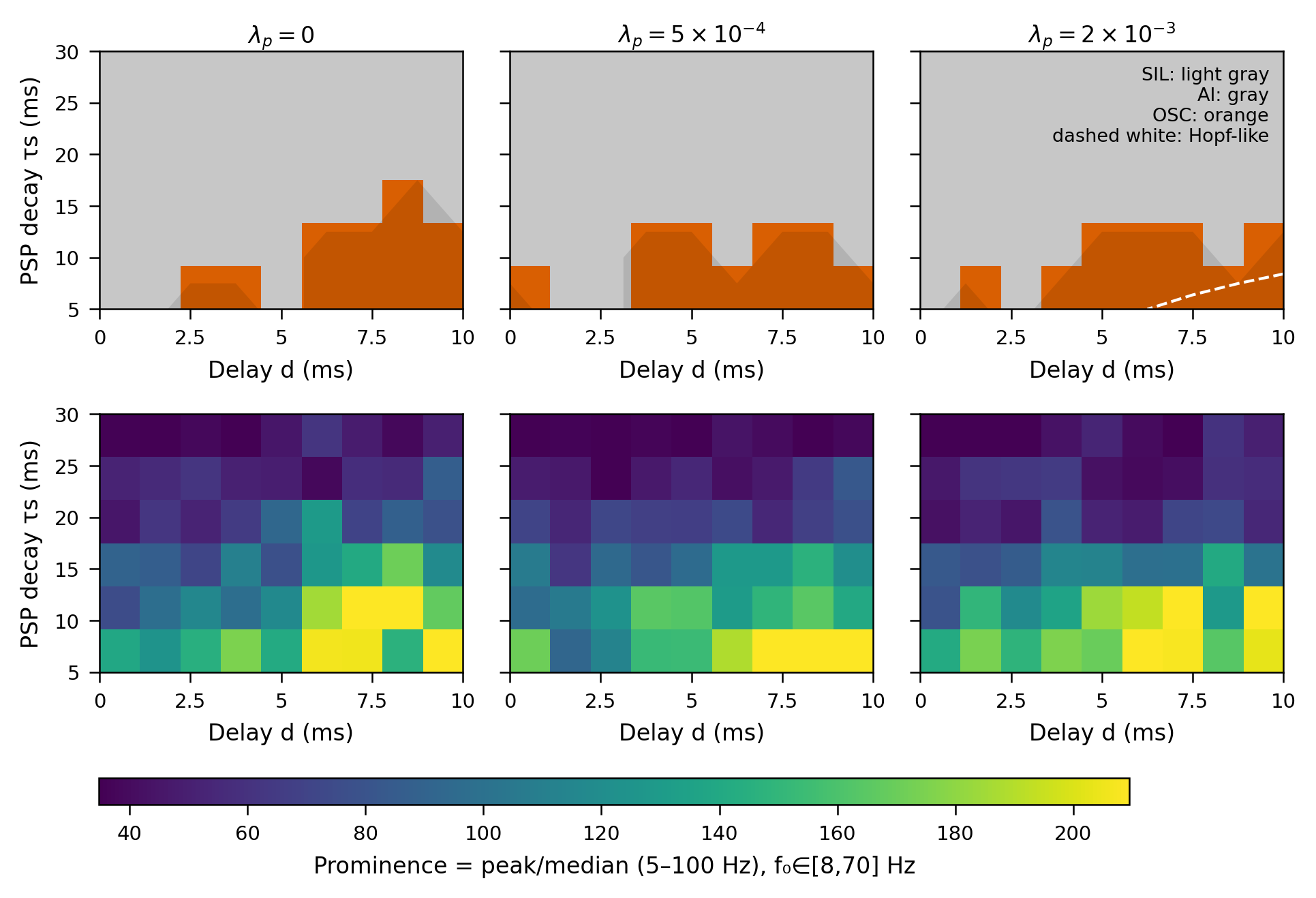}
  \vspace{-0.35em}
  \caption{
    \textbf{Regime maps of a balanced spiking network under PSP decay ($\tau_s$), delay ($d$), and plasticity rate ($\lambda_p$).} 
    %
    %
    For each \(\lambda_p\) slice, the top row shows the mean regime classification across five seeds (SIL, AI, and OSC), with the white dashed line indicating a coarse Hopf-reference boundary for the AI--OSC transition.  
    %
    %
    The bottom row shows the corresponding mean PSD prominence. The mapped parameter space spans \(\tau_s=5\)--30~ms and \(d=0\)--10~ms, providing an exploratory map of SIL--AI--OSC transitions.
    %
    %
    %
  }
  \label{fig:map}
\end{figure}

The resulting regime maps reveal how oscillatory states evolve across the \((\tau_s,d,\lambda_p)\) parameter space. Figure~\ref{fig:map} shows, for each \(\lambda_p\) slice, the regime classification across SIL, AI, and OSC states in the top row, together with the corresponding mean spectral prominence in the bottom row.
%
%
At \(\lambda_p=0\), networks were largely confined to asynchronous--irregular (AI) or silent (SIL) states.  
%
%
Increasing \(\lambda_p\) to \(2\times10^{-3}\) expanded the oscillatory (OSC) region toward shorter \(\tau_s\) and moderate--to--long delays.
%
%
The dashed Hopf boundary is included as a coarse theoretical reference for the observed oscillatory transition boundary in the mapped parameter space.
%
%

\begin{figure}[!t]
  \centering
  \vspace{-0.3em}
  \includegraphics[width=1.0\linewidth]{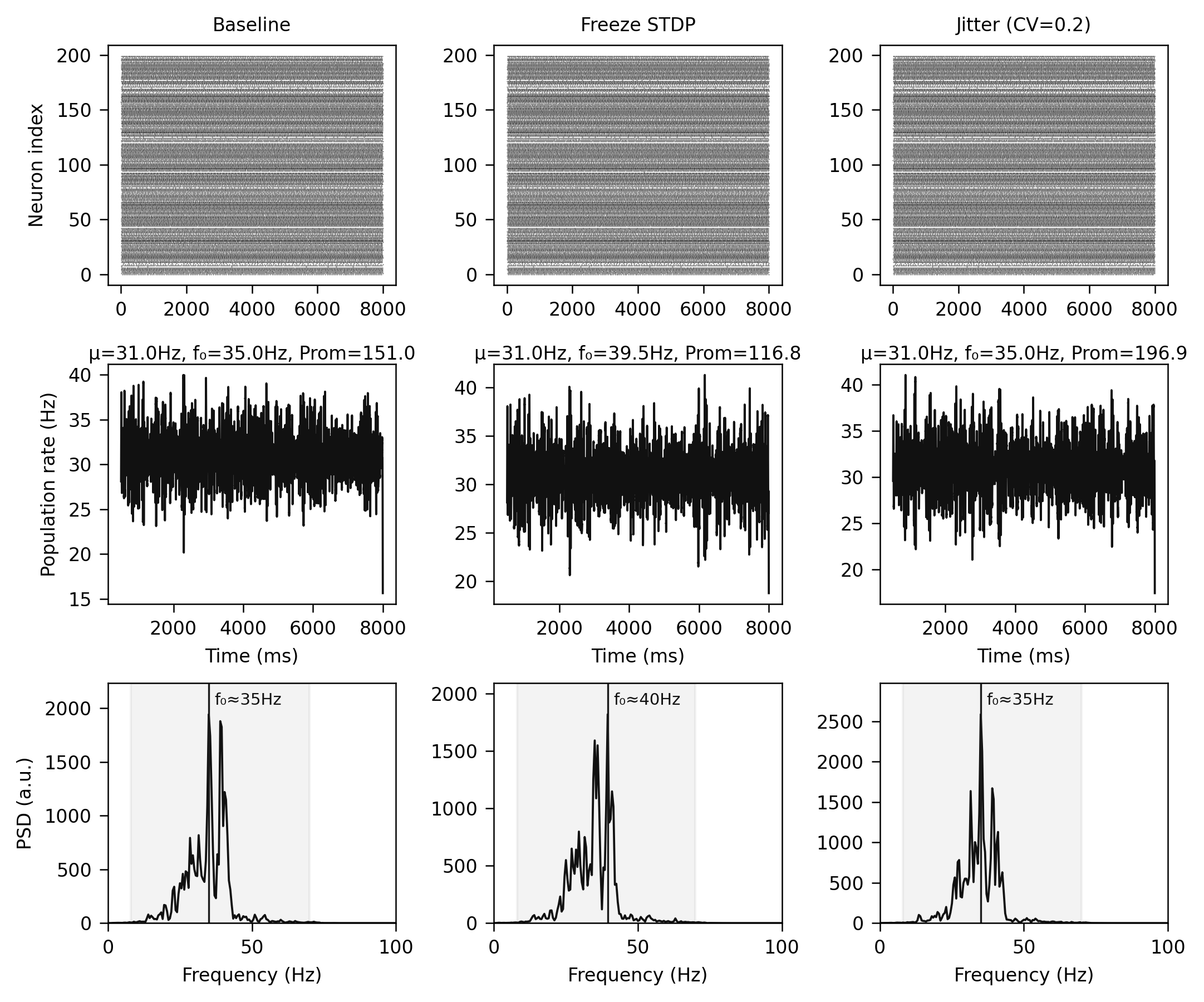}
  \vspace{-0.35em}
  \caption{\textbf{Control experiments in a representative oscillatory regime.}
  The operating point was selected from the highest-prominence region of the \(\lambda_p=2\times10^{-3}\) slice (\(\tau_s=5\)~ms, \(d=6.25\)~ms). The top row shows spike rasters under Baseline, Freeze, and Jitter conditions, the middle row shows population firing-rate traces, and the bottom row shows power spectra (5--100~Hz) with the dominant frequency (\(f_0\)) and spectral prominence (Prom).
  }
  \label{fig:controls}
\end{figure}

To further examine a representative oscillatory regime, control manipulations were applied at \(\lambda_p=2\times10^{-3}\), \(\tau_s=5\) ms, and \(d=6.25\) ms, as shown in Fig.~\ref{fig:controls}, with this exemplar selected from the highest-prominence region in the \(\lambda_p=2\times10^{-3}\) slice of Fig.~\ref{fig:map}.
%
%
Spike rasters (top) indicate comparable overall firing activity across Baseline (STDP active), Freeze (\(\lambda_p \to 0\)), and Jitter (\(\mathrm{CV}=0.2\)).  
%
%
Population-rate traces (middle) suggest slightly weaker rhythmic envelope fluctuations under Freeze, whereas delay jitter appears to enhance rhythmic modulation. 
This trend is further supported by the frequency-domain view, where the power spectra (bottom, 5--100~Hz) show a slight upward shift of the dominant frequency under Freeze and stronger rhythmic peaks under Jitter.
Quantitatively, Table~1 confirms that Freeze reduced Prom by 20.4\% (\(g=-0.93\)), whereas Jitter increased it by 50.8\% (\(g=+0.75\)), with only minor shifts in \(f_0\) and minimal change in mean firing rate.
%
%
%
This dissociation suggests that STDP and temporal delay variability primarily reshape rhythmic coherence rather than overall firing output within this representative oscillatory regime.
%
%

%
\begin{table}[!t]
  \centering
  \caption{\textbf{Control-condition summary across five seeds (mean $\pm$ SD).}
  Prom denotes the PSD peak-to-median ratio (5--100~Hz), $f_0$ the dominant frequency, Rate the mean firing rate, and $g$ Hedges' effect size measured relative to Baseline.
  }
  \vspace{0.15em}
  \renewcommand{\arraystretch}{1.0}
  \setlength{\tabcolsep}{4pt}
  \small
  \begin{adjustbox}{max width=\linewidth}
    \begin{tabular}{@{}lccccc@{}}
      \hline
      \textbf{Condition} & \textbf{Prom} & \textbf{$f_0$} & \textbf{Rate} & \textbf{Prom $\Delta$ (\%)} & \textbf{$g$} \\
      \hline
      Baseline & $175.2 \pm 41.8$ & $37.1 \pm 1.9$ & $31.0 \pm 0.8$ & --- & --- \\
      Freeze   & $139.4 \pm 13.8$ & $38.0 \pm 1.8$ & $31.0 \pm 0.8$ & $-20.4$ & $-0.93$ \\
      Jitter   & $264.2 \pm 129.9$ & $36.1 \pm 1.1$ & $31.0 \pm 0.8$ & $+50.8$ & $+0.75$ \\
      \hline
    \end{tabular}
  \end{adjustbox}
  \label{tab:stats}
\end{table}

This local control experiment provides a mechanistic illustration of a representative high-prominence point identified in the global regime maps of Fig.~\ref{fig:map}, suggesting how STDP and delay variability may locally shape rhythmic coherence within oscillatory regions.
%
%
%

\section*{Conclusion}
This study systematically mapped how postsynaptic decay, conduction delay, and plasticity rate jointly shape oscillatory states in balanced spiking networks. 
By combining recurrent spiking-network simulations with a coarse Hopf-reference boundary, we directly visualized SIL--AI--OSC transitions across the \((\tau_s,d,\lambda_p)\) parameter space.
%
%

As \(\lambda_p\) increased, the oscillatory region expanded toward shorter \(\tau_s\) and moderate-to-long delays, while the prominence maps further highlighted where rhythmic strength was strongest. Representative control analyses further show that STDP freezing weakens rhythmic coherence, whereas delay jitter enhances it, with only minor shifts in dominant oscillation frequency and minimal change in mean firing rate. Overall, these findings provide a clear global-to-local view of how multiple interacting time scales reshape oscillatory network dynamics.
%
%

Within this mapped network setting, the resulting regime maps reduce the need for repeated manual parameter trial-and-error and provide a more direct reference for locating oscillatory operating regions and analyzing rhythm formation. 
%

As a result, the mapped parameter space provides a useful reference for operating-point selection, synchrony modulation studies, and future biologically grounded spiking-network modeling within similar balanced-network settings.
%
%
%

{\small
\bibliographystyle{abbrv}
\bibliography{references}
}

\end{document}